\documentclass{segabs}
\usepackage{amsmath}
\usepackage{graphicx}
\usepackage{multicol}
\usepackage{cuted}
\usepackage{subfig}
\usepackage{booktabs}

\begin{document}

\title{Application of Machine Learning in Rock Facies Classification with Physics-Motivated Feature Augmentation}

\renewcommand{\thefootnote}{\fnsymbol{footnote}} 
\renewcommand{\figdir}{Fig} 

\author{Jie Chen$^1$, Yu Zeng$^2$$^{\dagger}$\\
\\
$^1$ jie.ch.2000@gmail.com   \\
$^2$ yu.zeng@alumni.duke.edu \\
$^{\dagger}$ Corresponding author \\
}

\righthead{Application of Machine Learning in Rock Facies Classification with Physics-Motivated Feature Augmentation}

\maketitle

\begin{abstract}
With recent progress in algorithms and the availability of massive amounts of computation power, application of machine learning 
techniques is becoming a hot topic in the oil and gas industry. One of the most promising aspects to apply machine learning to 
the upstream field is the rock facies classification in reservoir characterization, which is crucial in determining the net pay 
thickness of reservoirs, thus a definitive factor in drilling decision making process. For complex machine learning tasks like facies classification, 
feature engineering is often critical. This paper shows the inclusion of physics-motivated feature interaction in feature augmentation 
can further improve the capability of machine learning in rock facies classification. We demonstrate this approach with the SEG 2016 
machine learning contest \citep{tle_competition} dataset and the top winning algorithms \citep{xgboost}. The improvement is roboust and can be $\sim5\%$ 
better than current existing best F-1 score, where F-1 is an evaluation metric used to quantify average prediction accuracy.

\end{abstract}

\section{Introduction}
Machine learning, an application of Artificial Intelligence, is a very attractive alternative to traditional way of data analysis
by designing a system which can automatically learn from data and generalize from examples. Fuelled by the signifcant
advancement in computing power, machine learning is gaining more and more attention with many successful applications
including web search, spam filters, advertisement placement, stock trading, drug design, credit scoring, fraud detection,
etc.. 

Recently, machine learning is becoming a hot topic in the oil and gas field. Geoscientists are realizing the power of
machine learning techniques and are exploiting to apply them in various ways along the industry's production chains. To this regard, SEG hosted in 2016 a machine learning contest \citep{tle_competition} to tackle this problem.
The competition results in a collaborative technology outcome \citep{Paolo, Licheng}, utilizing the
power of AI algorithm and feature engineering, the latter is shown to be more effective way for
the success \citep{ml_contest_summary}.
Later after the contest, researchers from BP \citep{bp_paper} show the rock facies classification accuracy can be
further improved by a cognitive deep learning algorithm.
Inspired by Archie's equation \citep{archie} which describes a relationship between resistivity
and porosity for reservoir rocks, in this paper, we incorporate additional feature engineering to
the competition dataset, and obtained a $\sim5\%$ increase in F1-score \citep{scikit-learn}, a combined
precision and recall evaluation metric that is used in the contest.

In this paper, we will first review the facies classification problem, followed by a description of the proposed facies
classification algorithm. After that, we will discuss our feature engineering approach based on a better understanding
of well-log measurements using domain knowledge. Finally, we will demonstrate how our feature engineering approach can
further improve the prediction results on top of the SEG machine learning contest top winner's solution.

\section*{Facies Classification Problem Review}
In Oct. 2016, Hall initiated an SEG competition \citep{tle_competition} to classify rock facies using artifical intelligence approaches in {\it The Leading Edge} journal. In that journal article, a {\it Python}-based tutorial using Support Vector Machine (SVM) algorithm is adopted to demonstrate a typical machine learning workflow.
A well-log dataset named {\it training\_data.csv} was provided for usage. It contains 3232 entries with each entry consists of 5 wireline log measurements, 2 indicator variables, and measurement depth, for 8 wells in the Hugoton field of southwest Kansas \citep{Dubois}.

For each entry, a numeric label is also provided to classify the corresponding facies at a given depth into one of the 9 categories. See Table~\ref{tab:Facies} for description of those categories.
A well named \texttt{SHANKLE} is extracted from this dataset and saved as a blind test set for later classifier evaluation while the rest 7 wells are used for classifier training/testing purpose. In the paper,
an F-1 score of 0.43 is achieved when evaluating the trained model using the blind test set from well \texttt{SHANKLE} \citep{tle_competition}.

For the competition, a dataset called {\it facies\_vectors.csv} is provided on GitHub \citep{GitHub}. The relationship between {\it training\_data.csv} and {\it facies\_vectors.csv} is that the former is a subset of the latter, which contains additional wells and some incomplete entries like Photoelectric Effect (PE). Since the competition was closed in 2017 and the final
results based on F-1 score were evaluated by the host using another blind dataset which is, however, not accessible for us to use and to compare, we thus stick to the
accessible {\it training\_data.csv}, which has known facies labels, for all our tests and comparisons.

To obtain a basis as benchmark for comparison, we use the same publicly accessible dataset {\it training\_data.csv} and the same train/test/blind split as presented in {\it The Leading Edge} journal \citep{tle_competition}.
Applying the same solution from the SEG machine learning contest's top winner (LA Team \citep{LA_team}) to the blind test data \texttt{SHANKLE} we set aside, an F-1 score of 0.58 was obtained. Here the same solution means same
feature engineering, same data conditioning, same algorithm, and same model parameters \citep{LA_team}. The purpose of this article is to show that, by incorporting additional
feature enginnering using well log related domain knowledge, this F-1 score can be further improved to 0.61, representing a $\sim5\%$ improvement in the prediction accuracy.

\begin{table}[t]
{\begin{tabular}{@{}cc@{}} \toprule
\hline
Numeric label for facies & Description \\
 \hline
1 & Nomarine sandstone (SS) \\
2 & Nonmarine coarse siltstone (CSiS) \\
3 & Nonmarine fine siltstone (FSiS) \\
4 & Marine siltstone and shale (SiSh) \\
5 & Mudstone (MS) \\
6 & Wackestone (WS) \\
7 & Dolomite (D) \\
8 & Packstone-grainstone (PS) \\
9 & Phylloid-algal bafflestone (BS) \\ 
 \hline
\end{tabular}}
\caption{Description of facies labels \citep{tle_competition}}
\label{tab:Facies}
\end{table}

Besides apparent measurement depth, a good understanding of the meaning of the other 7 scalar attributes is needed if we want to improve on feature engineering side for this machine learning task. The 7 scalar attributes are:

\begin{itemize}
\item Gamma Ray (\texttt{GR}) measures $\gamma$-ray emissions from radiactive formations. Different formations will have differnt $\gamma$-ray signatures. Gamma Ray Logs can be used for correlation, shale content evaluation, and mineral analysis.
\item Resistivity (\texttt{ILD\_log10}) measures the ability of the subsurface materials to resist or inhibit electrical conduction. In our paper, it is re-labelled as $\log_{10}F$ for ease of dicussion.
\item Photoelectric effect (\texttt{PE}) measures the emission of electrons with incident photons.
\item Average neutron-density porosity (\texttt{PHIND}) measures a formation's porosity by analyzing neutron energy losses in porous formations. Neutron energy loss will occur most in the part of the formation that has the highest hydrogen concentration. In our paper, it is relabled as $\phi$ for ease of discussion.
\item Neutron-density porosity difference (\texttt{DeltaPHI}) provides porosity difference information based on neutron logs.
\item Nonmarine/marine indicator (\texttt{NM\_M}) is a binary indicator with value 1 and 2, indicating marine and non-marine facies, based on human interpretation.
\item Relative position (\texttt{RELPOS}) indicates the index position of each depth layer starting from 1 for the top layer. It is also based on human interpretation.
\end{itemize}

In machine learning language, this problem can be described as a small supervised, multiclass classification task whose goal is to find a model using the above features to predict the facies classification on previously unseen events. The evaluation metric to be used will be F-1 scoring, which combines both precision and recall and yields a single
measurement of accuracy.

Same as any machine learning problem, in order to obtain a higher F-1 score on blind dataset, efforts can be made in both algorithm side and feature engineering side.

In the area of machine learning algorithms, among many well-known ones, ensembles of decision trees, especially extreme gradient boosted trees, are often highly effective for multi-class
classification tasks. It turned out that, in the SEG machine learning competition, all of the top 10 models actually used an extreme gradient boosted tree package \texttt{XGBoost} \citep{xgboost}.
A brief description of \texttt{XGBoost} will be given in the next section. Other traditional methods, including SVM, $k$-NN, Multi-Layer Perceptron (MLP), often yield less impressive result.
The power of Deep Neural Networks (DNN) seems to be limited by this relatively small data set and is not quite effective for this task \citep{ml_contest_summary}.

Feature engineerging is another key part to successfully solve a complex machine learning problem. Domain-specific knowledge, sometimes, with intuition and insight, will play an
important role. It is not rare that features look irrelevant in isolation may become relevant when combined together. It turns out, for this facies classification task, proper
feature engineering can significantly boost the prediction accuracy.

\section{Facies Classification Algorithm}
The facies classification algorithm we used belongs to {\it ensembles of decision trees}, which combines prediction results from a large number of decision trees.
As demonstrated in many applications, methods based on {\it ensembles of decision trees} are in general effective and are well-suited for classification type of problems.
However, one limitation of these methods is they tend to overfit training data. A method called {\it gradient boosted trees} was proposed by Friedman \citep{friedman_boost}
to avoid this overfit tendency. A scalable end-to-end tree boosting algorithm system, called \texttt{XGBoost} \citep{xgboost}, implemented this idea in the package. Besides the regularized objective described in Eqn.\ref{eqn:obj},
\begin{equation}
  \mathcal{L}(\phi) = \sum_{i}L(\hat{y_{i}}, y_i) + \sum_{k}\Omega(f_k)
  \label{eqn:obj}
\end{equation}
where $L$ is the training loss fuction with $\hat{y_i}$ and $y_i$ to be prediction and target for $i$-th leaf, respectively, $\Omega$ is the regularization term, and $f_k$ represents an indepdent tree structure, \texttt{XGBoost} implements two additional techniques called shrinkage and column subsampling to further prevent overfitting. It has been widely used in data science community to solve challenging machine learning problems. Recent Kaggle competition and KDDCup competition winning results on
various topics show that about $\sim60\%$ of the winning solutions utilized \texttt{XGBoost} \citep{xgboost}.

\section{Benchmark Solution}

The final winner of this SEG competition is from LA Team, who used \texttt{XGBoost} algorithm plus a set of augmented features that were also used by all the top-ranked teams.
Table~\ref{tab:la_para} summarizes a list of LA Team's model parameters. The feature augmentation they adopted is actually originated from the third-place ISPL team, who later summarized
their result and published in 2017 SEG Annual Meeting \citep{Paolo}.

\begin{table*}[t]
\centering
{\begin{tabular}{@{}cccccc@{}} \toprule
\hline
Learning Rate & Max Depth & Min Child Weight & $N_{estimator}$ & Seed & Colsample$\_$bytree \\
 \hline
0.12 & 3 & 10 & 150 & 10 & 0.9 \\ 
 \hline
\end{tabular} }
\caption{\texttt{XGBoost} parameters from SEG competition top winner LA Team \citep{LA_team}.}
\label{tab:la_para}
\end{table*}

To construct our benchmark for future comparisons, we followed LA Team's solution and applied it to the public data {\it training\_data.csv}.
The same train/test split as presented in {\it The Leading Edge} journal is performed on this dataset, by saving well \texttt{SHANKLE}
as a blind evaluation set while the rest data used for training and testing, with 5\% of the data for the cross-validation and random state set to be 42.
A benchmark F-1 score of 0.58 was obtained using the \texttt{SHANKLE} blind evaluation set. This F-1 score of 0.58 is signifcant improvement over the baseline accuracy
of 0.42 as demonstrated by Hall in the geophysical tutorial using SVM \citep{tle_competition}.

To get an idea of the amount of improvement from algorithm side and feature engineering side, respectively, we performed two additional tests. In the first test, original raw
features without feature augmentation are used. SVM-based results and \texttt{XGBoost}-based results are compared. An improvement of 0.09 in F-1 score on \texttt{SHANKLE} blind evaluation set is obtained. On top of that result, feature augmentation is performed by exploiting the fact that spatial correlation exists in the data. A further improvement of 0.07 in F-1 score on \texttt{SHANKLE} blind evaluation set is obtained. These observations are consistent with the conclusions from other papers \citep{tle_competition, ml_contest_summary}.

The above tests show, beyond model choice from different algorithms, feature engineering is a key factor in achieving higher accuracy of the classification model. This also demonstrates that insights from human experts still have an important role in machine learning tasks even with significant advancement in algorithms.

\section{Physics Motivation}
For many maching learning applications, there is no doubt that the most important factor causing a project's success or failure is the features used.
Learning is usually easy with many independent features that each correlates well with the class. It is however difficult when the class turns out to be a very complex
function of the features where feature interaction has to be examined carefully \citep{friedman_ensemble}. Although the benchmark solution  expands the feature space 
via additional feature augmentation, including quadratic expansion, second
order feature interaction and feature derivatives, resulting in a significant F-1 score improvement, we believe physics based and domain 
knowledge based feature augmentation can provide additional information to separate facies and can further improve the prediction accuracy. 
This is the purpose of this article.

\section{Data Exploration}
To get an overall feeling, Fig. ~\ref{fig:measure_distribution} and Fig.~\ref{fig:facies_distribution} below show number of measurements at wells in {\it training\_data.csv} and overall distribution of facies in {\it training\_data.csv}, respectively. For each indiviual well, the distribution of facies can vary a lot. Correlations between scalar attributes are summarized in Table \ref{tab:corr2}, from which we can tell \texttt{ILD\_log10} and \texttt{PHIND}, for example, are highly anti-correlated.

\begin{figure}[!ht]
  \centering
  \includegraphics[scale=0.5]{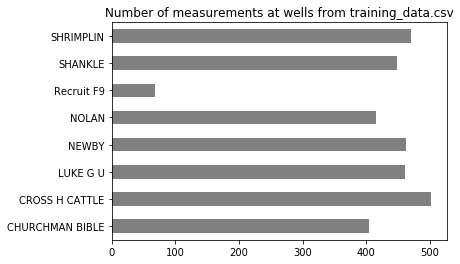}
  \caption{Number of facies measurements for each well in {\it training\_data.csv}}
  \label{fig:measure_distribution}
\end{figure}

\begin{figure}[!ht]
  \centering
  \includegraphics[scale=0.6]{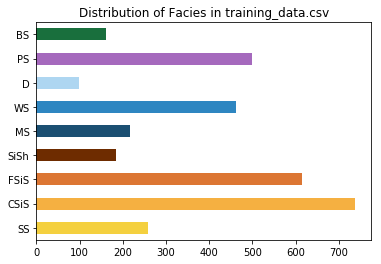}
  \caption{Distribuion of facies classes in {\it training\_data.csv}}
  \label{fig:facies_distribution}
\end{figure}

\begin{table*}[b]
\centering
{\begin{tabular}{@{}rrrrrrrrrr@{}}  \toprule
\hline
Attirbute & Facies & Depth & GR & ILD\_log10 & DeltaPHI & PHIND & PE & NM\_M & RELPOS \\
\hline
Facies       & 1.000 & 0.340 & -0.344 &  0.394 & -0.234 & -0.356 &  0.704 &  0.855 &  0.069 \\
Depth        &       & 1.000 & -0.064 &  0.178 & -0.091 & -0.074 &  0.278 &  0.297 &  0.001 \\
GR           &       &       &  1.000 & -0.156 &  0.190 &  0.248 & -0.289 & -0.281 & -0.173 \\
ILD\_log10   &       &       &        &  1.000 & -0.118 & -0.523 &  0.385 &  0.519 &  0.088 \\
DeltaPHI     &       &       &        &        &  1.000 & -0.250 &  0.011 & -0.174 &  0.037 \\
PHIND        &       &       &        &        &        &  1.000 & -0.573 & -0.488 & -0.035 \\
PE           &       &       &        &        &        &        &  1.000 &  0.657 &  0.019 \\
NM\_M        &       &       &        &        &        &        &        &  1.000 &  0.037 \\
RELPOS       &       &       &        &        &        &        &        &        &  1.000 \\
\hline
\end{tabular}}
\caption{Correlation coefficients between scalar attributes}
\label{tab:corr2}
\end{table*}

\subsection{Feature examination: Resistivity}
Resistivity describes the resistance of a material to the flow of electric current. It can be described by Eqn.~\ref{eqn:resistivity} as\
 follows:
\begin{equation}
  R = \frac{r\cdot A}{L} = \frac{V\cdot A}{I \cdot L}
  \label{eqn:resistivity}
\end{equation}
where R is resistivity of sample in unit of $\Omega m$, $r$ is resistance in unit of $\Omega$, $A$ is cross-section area of sample in un\
it of $m^2$ and $L$ is length of sample in unit of $m$.

A formation's true resistivity $R_t$ is its resistivity when not contaminated by drilling fluids.
It may contain formation water only or formation water and hydrocarbons. It is fundamental to use
a valid $R_t$ for the presence of hydrocarbons when analyzing well log measurements.
With the presence of drilling fluid's invasion, resistivity of mud filtrate $R_{mf}$ can affect $R_t$ measurement.
The measured resistivity can be either greater than, less than, or equal to $R_t$ and can distort
deep resistivities. To get a valid $R_t$ value, corrections must be made to attenuate such distortions.

\begin{figure}[!ht]
  \centering
  \includegraphics[scale=0.5]{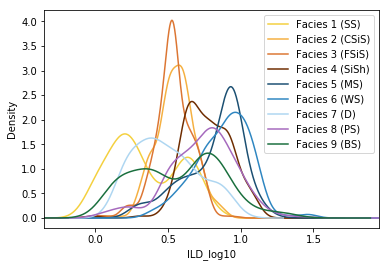}
  \caption{Distribution of $log_{10}F$ by facies}
  \label{fig:logF_qc}
\end{figure}

Fig.~\ref{fig:logF_qc} shows drilling fluid corrected distribution of formation resisitivity $F$ in logrithmetic form $log_{10}F$
(attribute \texttt{ILD\_log10}).

\subsection{Feature examination: Neutron Density Porosity}
Neutron logs are used to measure the hydrogen content in a formation.
The emitted neutrons from a chemical source will collide with nuclei in the formation and thus lose energy.
With enough collisions, a neutron will be absorbed by the formation and a $\gamma$-ray will be emitted. In
neutron-hydrogen collisions, the average energy transfers to the hydrogen nucleus is about $\sim50\%$ of the
energy of the original neutron since the mass of a neutron is close to the mass of a hydrogen (proton).
This indicates that materials with large hydrogen content will slow down neutrons.

\begin{figure}[!ht]
  \centering
  \includegraphics[scale=0.5]{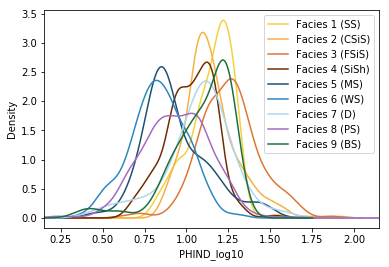}
  \caption{Distribution of $log_{10}\phi$ by facies}
  \label{fig:logPhi_qc}
\end{figure}

In a porous formation, hydrogen contents tend to concentrate in the fluid-filled pores, formation's porosity
can thus by inferred by measuring neutron energy losses.
Fig.~\ref{fig:logPhi_qc} shows distribution of neutron density porosity $\phi$ in logrithmetic form $log_{10}\phi$ (attribute \texttt{PHIND\_log10}).

\subsection{Relationship between Resistivity and Porosity}
In the famous paper on classification of carbonate reservoir rocks and petrophysical considerations \citep{archie}, Archie proposed an empirical formula to calculate the water saturation ($S_{w}$) in a formation next to a borehole from well log parameters.
This formula, known as Archie equation, is shown in Eqn.~\ref{eqn:archie1}  below:
\begin{equation}
  F = \frac{R_0}{R_w} = C\cdot\phi^{-m}
  \label{eqn:archie1}
\end{equation}
where $F$ is called formation resistivity, $R_w$ with unit $\Omega m$ is the resistivity of reservoir water, $R_0$ is the resistivity of reservoir rock saturated with reservoir water, and $R_t$ is the resistitivy of reservoir rock saturated with oil and water, $C$ is tortuousity constant, $m$ is cementation factor, which depends on rock formation.
After taking the logrithmetic operation on both sides, Eqn.~\ref{eqn:archie1} can be re-written as:
\begin{equation}
  log_{10} F = log_{10}C - m\cdot log_{10}\phi
  \label{eqn:archie2}
\end{equation}
Archie's equation in the form of Eqn.~\ref{eqn:archie2}, though empirical, suggests there exists a linear relationship between $log_{10}F$ and $log_{10}\phi$ which may vary for different rock types. This information, when provided, may further improve
the accuracy of machine learning classification tasks.

\begin{figure}[!ht]
  \centering
  \includegraphics[scale=0.5]{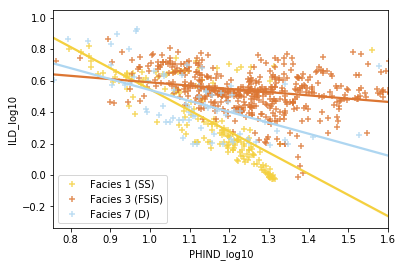}
  \caption{$log_{10}F$ vs. $log_{10}\phi$ scatter plot overlaid with linear regression lines}
  \label{fig:logF_vs_logPhi}
\end{figure}

For better display and without loss of generality, three facies (facies 1, 3, and 7) are chosen to generate $log_{10}F$ vs. $log_{10}\phi$ plot as shown in Fig.~\ref{fig:logF_vs_logPhi}, with simple linear regression lines overlaid on top.
Fig.~\ref{fig:logF_vs_logPhi} suggests the ratio between logrithmetic
resistivity $log_{10}F$ and logrithmetic neutron-density porosity $log_{10}\phi$ have good discrimination power for different facies. Fig. ~\ref{fig:logF_vs_logPhi} also suggests the $log_{10}F$ vs. $log_{10}\phi$ discrimination power will not be the same for different facies pairs. For example, $log_{10}F$ vs. $log_{10}\phi$ can better discriminate facies 1 and facies 3 than facies 1 and facies 7.

\section{Classifier Evaluation}
Starting with benchmark, with additional feature augmentation using $\frac{log_{10}F}{log_{10}\phi}$, the final averaged F-1 score of 0.61 is achieved. This is a $\sim 5\%$ improvement over the benchmark F-1 score of 0.58 under the same settings.

\begin{figure}[!ht]
  \centering
  \subfloat[from benchmark approach]{{\includegraphics[scale=0.5]{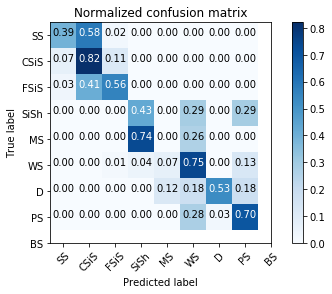}} }%
  \qquad
  \subfloat[from our approach]{{\includegraphics[scale=0.5]{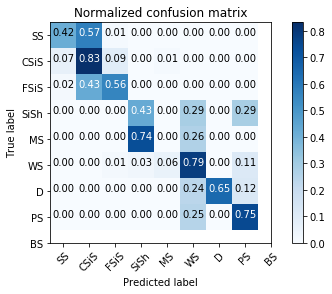}} }%
  \caption{Comparison of normalized confusion matrix: benchmark approach vs. our approach}
  \label{fig:conf_matrix_comp}
\end{figure}

A confusion matrix plot like Fig.~\ref{fig:conf_matrix_comp}a is used to give a more direct visual display of prediction accuracy for each facies class. In an ideal case using a perfect classifier, the confusion matrix should become an identity matrix $I_{n\times n}$, with all diagonal elements to be 1 and off-diagonal elements to be 0. The comparison of confusion matrix in Fig.~\ref{fig:conf_matrix_comp}a and in Fig.~\ref{fig:conf_matrix_comp}b clearly shows that confusion matrix
from our solution is overall more focused around diagonal elements while reducing the values in off-diagonal elements, indicating an overal improvement in prediction accuracy.

\begin{figure*}[t]
  \centering
  \includegraphics[width=\textwidth]{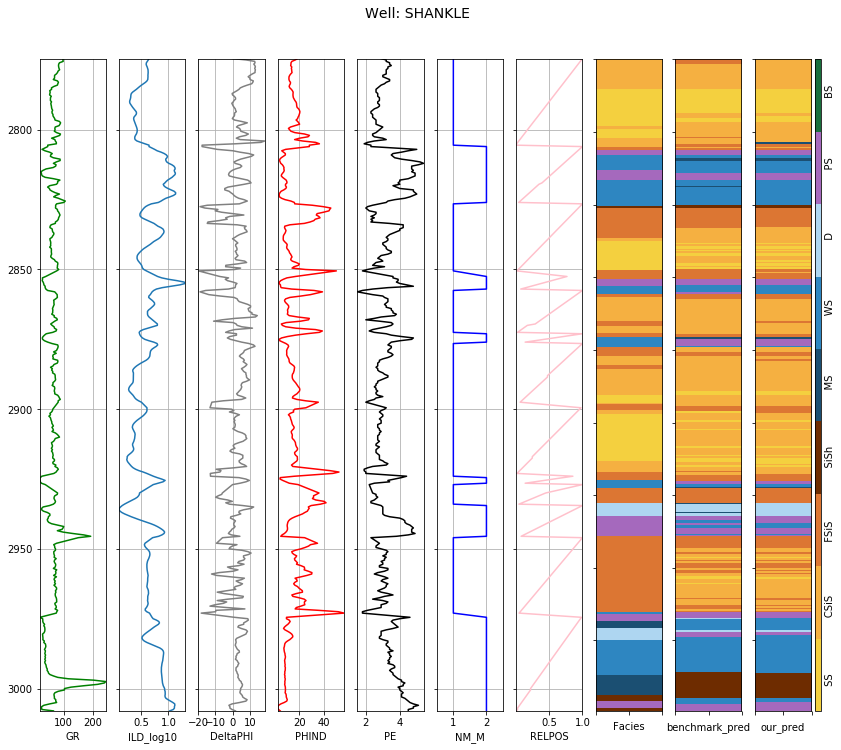}
  \caption{Wireline log measurements and comparison of facies classfication between true values, benchmark predictions and our predictions.}
  \label{fig:logdisplay}
\end{figure*}

Fig.~\ref{fig:logdisplay} shows a more obvious comparisons of true facies, predicted facies from benchmark approach and predicted facies from our approach. The 7 raw features are plotted on the left of the figure, while the true and predicted facies categories are plotted on the right, using the same depth information. Overall, our approach provides a better facies classification
prediction than benchmark.

To verify the robustness of our approach, a few sanity checks are performed.
To eliminate the possible bias introduced by the chosen random seed, we tested our approach
using different random seeds and find the resulting improvement on F-1 score is robust and 
is consistently at $2\% \sim 7\%$ level. We also performed 7-fold cross-validation test by saving one of the 8 wells as blind well for test, and found the F-1 score
is also improved at $\sim 5\%$ level consistently. As another cross check, with the same train/test/blind split
and the same random state, we repeated the exactly identical SVM-based workflow as demonstrated in the SEG machine learning contest paper \citep{tle_competition} by adding new feature $\frac{ log_{10} F}{ log_{10} \phi}$ (i.e. \texttt{ILD\_log10}/\texttt{PHIND\_log10}), and the resulting F-1 score on blind evaluation dataset is increased from 0.43 to 0.44, which is $2.3\%$ improvement. These tests confirmed the uplift from our feature augmentation is robust against algorithm change, data sampling variation, and random seed initialization.
Though better than benchmark with robust improvement, the model we obtained is still not perfect. As can be seen in Fig.~\ref{fig:conf_matrix_comp} and Fig.~\ref{fig:logdisplay}, our model has trouble in making correct predictions for some facies, especifally those may blend into one another. A good example would be facies 4 (Marine siltstone and shale: SiSh) and facies 5 (Mudstone: MS). We expect further work on deep learning strategies for feature learning and inclusions of additional physical measurements can further improve the prediction accuracy.

\section{Conclusion and Discussion}
In this paper, we demonstrated feature augmentation incorporating domain knowledge can further improve the capability of machine learning. Motivated by Archie's equation and with the inclusion of additional feature created from resistivity and porosity measurements, robust improvement in F-1 score is obtained and can be $\sim5\%$ better over current existing best F-1 score. This example also shows there is ultimately no replacement for the insights human beings can put into feature engineering.
 
Currently, the dataset we use only contains limited number of physical attributes. To further improve in future rock facies classification tasks, additional types of well log measurements, including $V_p$, $V_s$, density etc., if provided, will probably further enhance machine learning's capability in achieving better accuracy. For example, $V_p/V_s$ is directly related to the Poisson's ratio \citep{Gercek}, which measures the ratio of lateral strain to axial strain in a rock and can be served as another distinctive discriminator in facies classification. 

For machine learning algorithms, the power of DNN seems to be limited by this relatively small dataset \citep{ml_contest_summary}, leading to worse prediction accuracy when compared with XGBoost. Recently, a noval stochastic gradient descent method is proposed to overcome the requirement of large samples in traditional DNN \citep{wang}. This would potentially improve the DNN-based F-1 score.

\section{ACKNOWLEDGMENTS}
The authors would like to thank the organizers of the SEG machine learning contest for making contest dataset and solution publicly available for further R\&D.

\onecolumn

\bibliographystyle{seg}  


\end{document}